\documentclass[10pt,twocolumn,letterpaper]{article}

\usepackage{cvpr}              

\usepackage{graphicx}
\usepackage{amsmath}
\usepackage{amssymb}
\usepackage{booktabs}
\usepackage[ruled,lined,linesnumbered]{algorithm2e}

\usepackage{color, colortbl}
\usepackage[dvipsnames]{xcolor}
\definecolor{citecolor}{HTML}{0071bc}
\usepackage[pagebackref,breaklinks=true,letterpaper=true,colorlinks,citecolor=citecolor,bookmarks=false]{hyperref}

\usepackage[capitalize]{cleveref}
\crefname{section}{Sec.}{Secs.}
\Crefname{section}{Section}{Sections}
\Crefname{table}{Table}{Tables}
\crefname{table}{Tab.}{Tabs.}

\def\cvprPaperID{***}
\def\confName{CVPR}
\def\confYear{2022}

\usepackage{comment}
\usepackage[colorlinks]{hyperref}
\usepackage{url}
\usepackage{amsmath,amssymb} 
\usepackage{times}
\usepackage{epsfig}
\usepackage{graphicx}
\usepackage{amsmath}
\usepackage{amssymb}
\usepackage{caption}
\usepackage{subcaption}
\usepackage{dblfloatfix}
\usepackage{setspace}
\usepackage{bbm}
\usepackage{multirow}

\renewcommand{\paragraph}[1]{\noindent\textbf{#1}}
\renewcommand{\vec}{\boldsymbol}
\usepackage{gensymb}
\usepackage{booktabs}
\newcommand{\T}{\mathcal{T}}
\let\oldsubsection\subsection
\renewcommand{\subsection}[1]{\vspace{-1mm}\oldsubsection{#1}\vspace{-1mm}}

\newcommand{\reffig}[1]{Figure~\ref{fig:#1}}

\newcommand{\refsec}[1]{Section~\ref{sec:#1}}

\newcommand{\reftbl}[1]{Table~\ref{tbl:#1}}

\newcommand{\refeqshort}[1]{\eqref{eq:#1}}

\newcommand{\lblsec}[1]{\label{sec:#1}}
\newcommand{\lbleq}[1]{\label{eq:#1}}

\newcommand{\lbltbl}[1]{\label{tbl:#1}}

\usepackage{hyperref}

\newcommand{\rowNumber}[1]{\textcolor{Cerulean}{#1}}

\begin{document}

\title{Simple Multi-dataset Detection}

\author{
  Xingyi Zhou$^{1}$ \quad Vladlen Koltun$^2$ \quad Philipp Kr{\"a}henb{\"u}hl$^1$\\
{$^1$The University of Texas at Austin \quad \quad $^2$Apple}
}

\maketitle

\begin{abstract}
How do we build a general and broad object detection system? We use all labels of all concepts ever annotated. These labels span diverse datasets with potentially inconsistent taxonomies. In this paper, we present a simple method for training a unified detector on multiple large-scale datasets. We use dataset-specific training protocols and losses, but share a common detection architecture with dataset-specific outputs. We show how to automatically integrate these dataset-specific outputs into a common semantic taxonomy. In contrast to prior work, our approach does not require manual taxonomy reconciliation. Experiments show our learned taxonomy outperforms a expert-designed taxonomy in all datasets. Our multi-dataset detector performs as well as dataset-specific models on each training domain, and can generalize to new unseen dataset without fine-tuning on them. 
Code is available at \url{https://github.com/xingyizhou/UniDet}.
\end{abstract}

\section{Introduction}
Computer vision aims to produce broad, general-purpose perception systems that work in the wild.
Yet object detection is fragmented into datasets~\cite{lin2014microsoft, MVD2017, shao2019objects365, OpenImages} and our models are locked into the corresponding domains.
This fragmentation brought rapid progress in object detection~\cite{ren2015faster,ge2021yolox,zhang2020varifocalnet,wang2020scaled,li2020generalizedv2,dai2021dynamic} and instance segmentation~\cite{he2017mask},
but comes with a drawback.
Single datasets are limited in both image domains and label vocabularies and do not yield general-purpose recognition systems.
Can we alleviate these limitations by unifying diverse detection datasets?

In this paper, we first make training an object detector on a collection of disparate datasets as straightforward as training on a single one.
Different datasets are usually trained under different training losses, data sampling strategies, and schedules. 
We show that we can train a single detector with separate outputs for each dataset, and apply dataset-specific supervision to each.
Our training mimics training parallel dataset-specific models with a common network.
As a result, our single detector takes full advantages of all training data,
performs well on training domains, and generalizes better to new unseen domains.
However, this detector produces duplicate outputs for classes that occur in multiple datasets.

\begin{figure}[t]
\centering
   \includegraphics[width=0.95\linewidth]{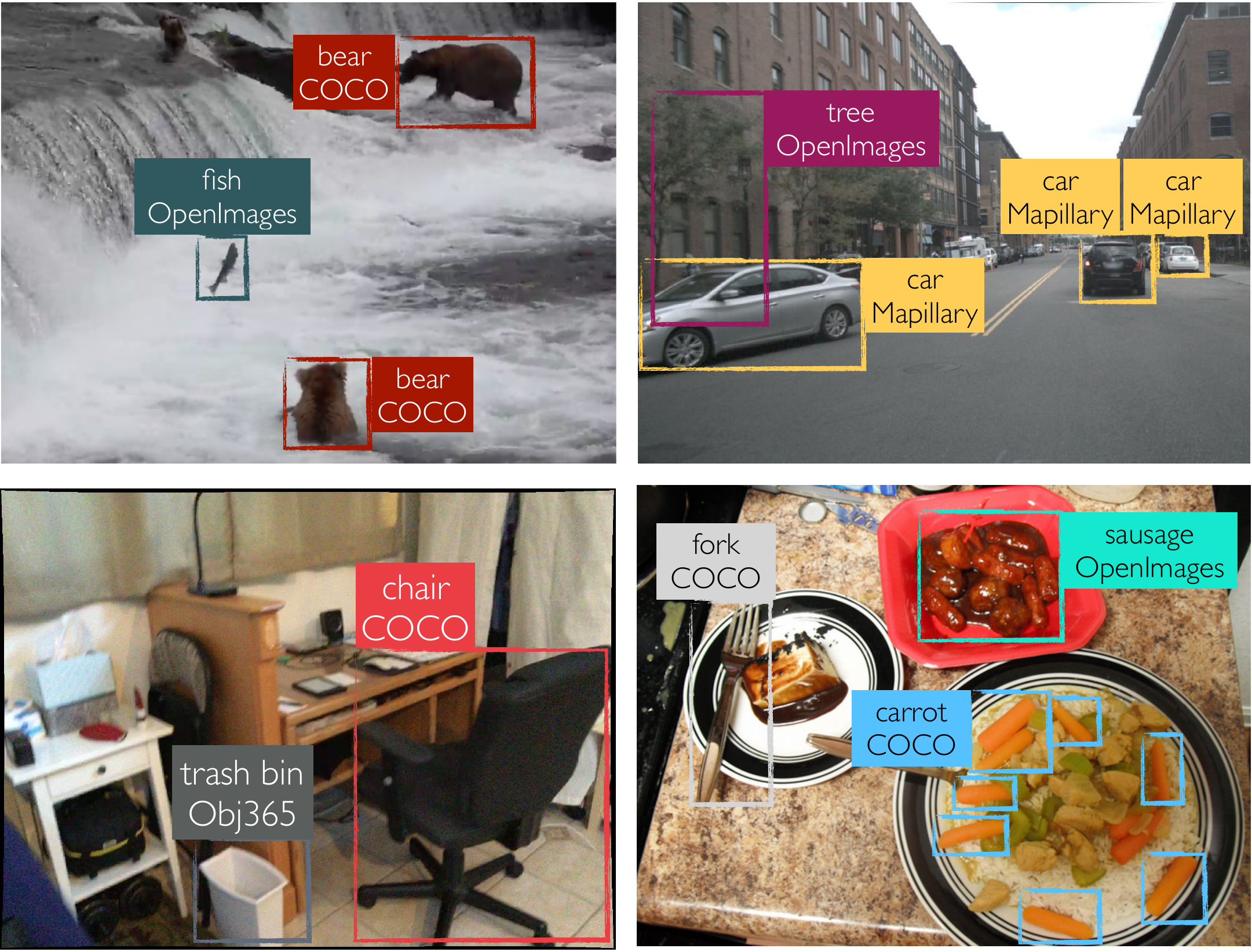}
   \vspace{-2mm}
   \caption{Different datasets span diverse semantic and visual domains. We learn to unify the label spaces of multiple datasets and train a single object detector that generalizes across datasets.}
\label{fig:teaser}
\vspace{-5mm}
\end{figure}

A core challenge is integrating different datasets into a common taxonomy, and training a detector that reasons about general objects instead of dataset-specific classes.
Traditional approaches create this taxonomy by hand~\cite{MSeg_2020_CVPR, Zhao_UniDet_ECCV20}, which is both time-consuming and error-prone.
We present a fully automatic way to unify the output space of a multi-dataset detection system using visual data only.
We use the fact that object detectors for similar concepts from different datasets fire on similar novel objects.
This allows us to define the cost of merging concepts across datasets, and optimize for a common taxonomy fully automatically.
Our optimization jointly finds a unified taxonomy, a mapping from this taxonomy to each dataset, and a detector over the unified taxonomy using a novel 0-1 integer programming formulation.
An object detector trained on this unified taxonomy has a large, automatically constructed vocabulary of concepts from all training datasets.

We evaluate our unified object detector at an unprecedented scale.
We train a unified detector on 3 large and diverse datasets:
COCO~\cite{lin2014microsoft}, Objects365~\cite{shao2019objects365}, and OpenImages~\cite{OpenImages}.
For the first time, we show that a single detector performs as well as dataset-specific models on each individual dataset.
A unified taxonomy further improves this detector.
Crucially, we show that models trained on diverse training sets generalize to new domains \emph{without retraining}, and outperform single-dataset models.

\section{Related Work}

\par \noindent \textbf{Training on multiple datasets.} In recent years, training on multiple diverse datasets has emerged as an effective tool to improve model robustness for depth estimation~\cite{Ranftl2020}, stereo matching~\cite{Yang_2019_CVPR}, and person detection~\cite{Hasan_2021_CVPR}.
In these domains, unifying the output space involves modeling different camera transformations or depth ambiguities.
In contrast, for recognition, dataset unification involves merging different \emph{semantic} concepts.
MSeg\cite{MSeg_2020_CVPR} manually unified the taxonomies of 7 semantic segmentation datasets and used Amazon Mechanical Turk to resolve inconsistent annotations between datasets.
In contrast, we propose to learn a label space from visual data automatically, without requiring any manual effort.

Wang et al. \cite{wang2019towards} train a universal object detector on multiple datasets, and gain robustness by joining diverse sources of supervision.
This is similar to our partitioned detector, while they work on small datasets and didn't model the training differences between different datasets.
Universal-RCNN~\cite{xu2020universal} trains an partitioned detector on
three large datasets~\cite{lin2014microsoft,krishna2017visual,zhou2017scene} and models the class relations with a inter-dataset attention module.
However again they use the same training recipe for all datasets, and produce duplicated outputs for the same object if it occurs in more one dataset.
Both Wang et al. \cite{wang2019towards} and MSeg\cite{MSeg_2020_CVPR} observe a performance drop in a single unified model.
With our dedicated training framework, this is not the case: our unified model performs as well as single-dataset models on the training datasets.
Also, these multi-headed models produce a dataset-specific prediction for each input image.
When evaluated in-domain, they require knowledge of the test domain.
When evaluated out-of-domain, they produce multiple outputs for a single concept.
This limits their generality and usability.
Our approach, on the other hand, 
unifies visual concepts in a single label space and yields a single consistent model that does not require knowledge of the test domain and can be deployed cleanly in new domains.

Zhao et al. \cite{Zhao_UniDet_ECCV20} trains a universal detector
on multiple datasets: COCO~\cite{lin2014microsoft}, Pascal VOC~\cite{egwwz-pvocc-10}, and SUN-RGBD~\cite{song2015sun}, with under 100 classes in total.
They manually merge the taxonomies and then train with cross-dataset pseudo-labels generated by dataset-specific models.
The pseudo-label idea is complementary to our work.
Our unified label space learning removes the manual labor, and works on a much larger scale:
we unify COCO, Objects365, and OpenImages, with more complex label spaces and $900+$ classes. 
YOLO9000~\cite{redmon2017yolo9000} combines detection and classification datasets to expand the detection vocabulary. 
LVIS~\cite{gupta2019lvis} extents COCO annotations to $>\!1000$ classes in a federated way.
Our approach of fusing multiple annotated datasets is complementary and can be operationalized with no manual effort to unify disparate object detection datasets.

\par \noindent \textbf{Zero-shot classification and detection} reasons about novel object categories outside the training set~\cite{fu2018recent, bansal2018zero}.
This is often realized by representing a novel class by a semantic embedding~\cite{norouzi2013zero} or auxiliary attribute annotations~\cite{farhadi2009describing}.
In zero-shot detection, Bansal et al. \cite{bansal2018zero} proposed a statically assigned background model to avoid novel classes being detected as background. Rahman et al. \cite{rahman2019transductive} used test-time training to progressively generate new class labels based on word embeddings. Li et al.\cite{li2019zero} leveraged external text descriptions for novel objects.
Our program is complementary: we aim to build a sufficiently large label space by merging diverse datasets during training, such that the trained detector transfers well across domains even without machinery such as word embeddings or attributes. 
Such machinery can be added, if desired, to further expand our model's vocabulary.

\section{Preliminaries}

Object detection aims to predict a location $b_i \in \mathbb{R}^4$ and a class-wise detection score $d_i \in \mathbb{R}^{|L|}$ for each object $i$ in image $I$.
The detection score describes the confidence that a bounding box belongs to an object with label $c \in L$, where $L$ is the set of all classes (label space) of the dataset $\mathcal{D}$.

Many existing works on object detection focus on the COCO dataset~\cite{lin2014microsoft}, which contains balanced annotations for 80 common object classes.
This class balance simplifies training and yields good generalization.
Training an object detector on COCO follows a simple recipe:
Minimize a loss $\ell$, usually box-level log-likelihood, over an sampled image $\hat I$ and its corresponding annotated bounding boxes annotations $\hat B$ from the dataset $\mathcal{D}$:
\begin{equation}
    \min_{\Theta} \mathbb{E}_{(\hat I, \hat B) \sim \mathcal{D}} \left[\ell(\mathcal{M}(\hat I; \Theta), \hat B)\right].\lbleq{coco_loss}
\end{equation}
Here, $\hat  B$ contains class-specific 
box annotations.
The loss $\ell$ operates on sets of outputs and annotations, and matches them using an overlap criterion.

Let's now consider training a detector on multiple datasets $\mathcal{D}_1, \mathcal{D}_2,\ldots$, each with their own label space $L_1, L_2, \ldots$.
A natural way to train on multiple datasets is to simply combine all annotations of all datasets into a much larger dataset $\mathcal{D} = \mathcal{D}_1 \cup  \mathcal{D}_2 \cup \ldots$, and merge their label spaces $L=L_1 \cup L_2 \cup \ldots$.
Labels that repeat across datasets are merged.
We then optimize the same loss with more data:
\begin{equation}
    \min_{\Theta} \mathbb{E}_{(\hat I, \hat B) \sim \mathcal{D}_1 \cup  \mathcal{D}_2 \cup \ldots} \left[\ell(\mathcal{M}(\hat I; \Theta), \hat B)\right].\lbleq{joint_loss}
\end{equation}
This has shown promise on smaller, evenly distributed datasets~\cite{wang2019towards,wu2019detectron2,egwwz-pvocc-10}.
It has the advantage that shared classes between the datasets train on a larger set of annotations.
However, modern large-scale detection datasets feature more natural class distributions that are imbalanced.
Objects365~\cite{shao2019objects365} contains $5\times$ more images than COCO and OpenImages~\cite{OpenImages} is $18\times$ larger than COCO. 
While the top $20\%$ of classes in Objects365 and OpenImages contain $19\times$ and $20\times$ more images than COCO, respectively, the bottom $20\%$ classes actually have fewer images than COCO.
This imbalance in class distributions and dataset sizes all but guarantees that a simple concatenation of datasets will not work.
In fact, not even the same loss \refeqshort{coco_loss} works for all datasets.
Most successful Objects365 models~\cite{gao2019objects365} employ class-aware sampling~\cite{shen2016relay}.
OpenImages models treat rare classes differently~\cite{tan2020equalization} and model the hierarchy of classes in the loss~\cite{peng2020large}.

This suggests that training a detector $M_k$ on a dataset $D_k$ requires a dataset-specific loss $\ell_k$:
\begin{equation}
    \min_{\Theta} \mathbb{E}_{(\hat I, \hat B) \sim \mathcal{D}_k} \left[\ell_k(\mathcal{M}_k(\hat I; \Theta), \hat B)\right].\lbleq{specific_loss}
\end{equation}
No single loss generalizes to all datasets.
In the next section, we present a different view of multi-dataset training and show how to train a model that performs well on all datasets.

\section{Training a multi-dataset detector}

\lblsec{splithead}

Our goal is to train a single detector $\mathcal{M}$ on $K$ datasets $\mathcal{D}_1, \ldots, \mathcal{D}_K$ with label spaces $L_1, \ldots, L_K$, and dataset-specific training objectives $\ell_1,\ldots, \ell_K$.
Our core insight is that we can train a unified detector in the same way as we train multiple dataset-specific detectors separately,
as long as we do not attempt to merge label spaces between different datasets.
This can be considered training
 $K$ dataset-specific detectors $\mathcal{M}_1, \ldots, \mathcal{M}_K$ in parallel, while \emph{sharing their backbone architecture $\mathcal{M}$}.
Each dataset-specific architecture shares all but the last layer with the common backbone.
Each dataset uses its own classification layer at the end.
We call this a {\bf partitioned detector} (Figure~\ref{fig:overview_b}).
We train a partitioned detector over all datasets by minimizing the $K$ dataset-specific losses:
\begin{equation}
    \min_{\Theta} \mathbb{E}_{\mathcal{D}_k}\left[\mathbb{E}_{(\hat I, \hat B) \sim \mathcal{D}_k}\left[ \ell_k(\mathcal{M}_k(\hat I; \Theta), \hat B)\right]\right].\lbleq{combined_loss}
\end{equation}
Here, evenly sampling datasets, i.e. showing the partitioned detector the same number of images from each dataset, works best empirically, as we will show in \refsec{results}.

While the partitioned detector learns to detect all classes, it still produces different dataset-specific outputs.
For example, it predicts a COCO-person separately from an Objects365-Person, etc.
Next we show how to convert this partitioned model into a joint detector that reasons about a \emph{unified} set of output labels $L = L_1\cup L_2 \cup\ldots$.

\subsection{Learning a unified label space}

\begin{figure*}
    \centering
    \begin{subfigure}{0.24\textwidth}
    \includegraphics[page=2, width=\linewidth]{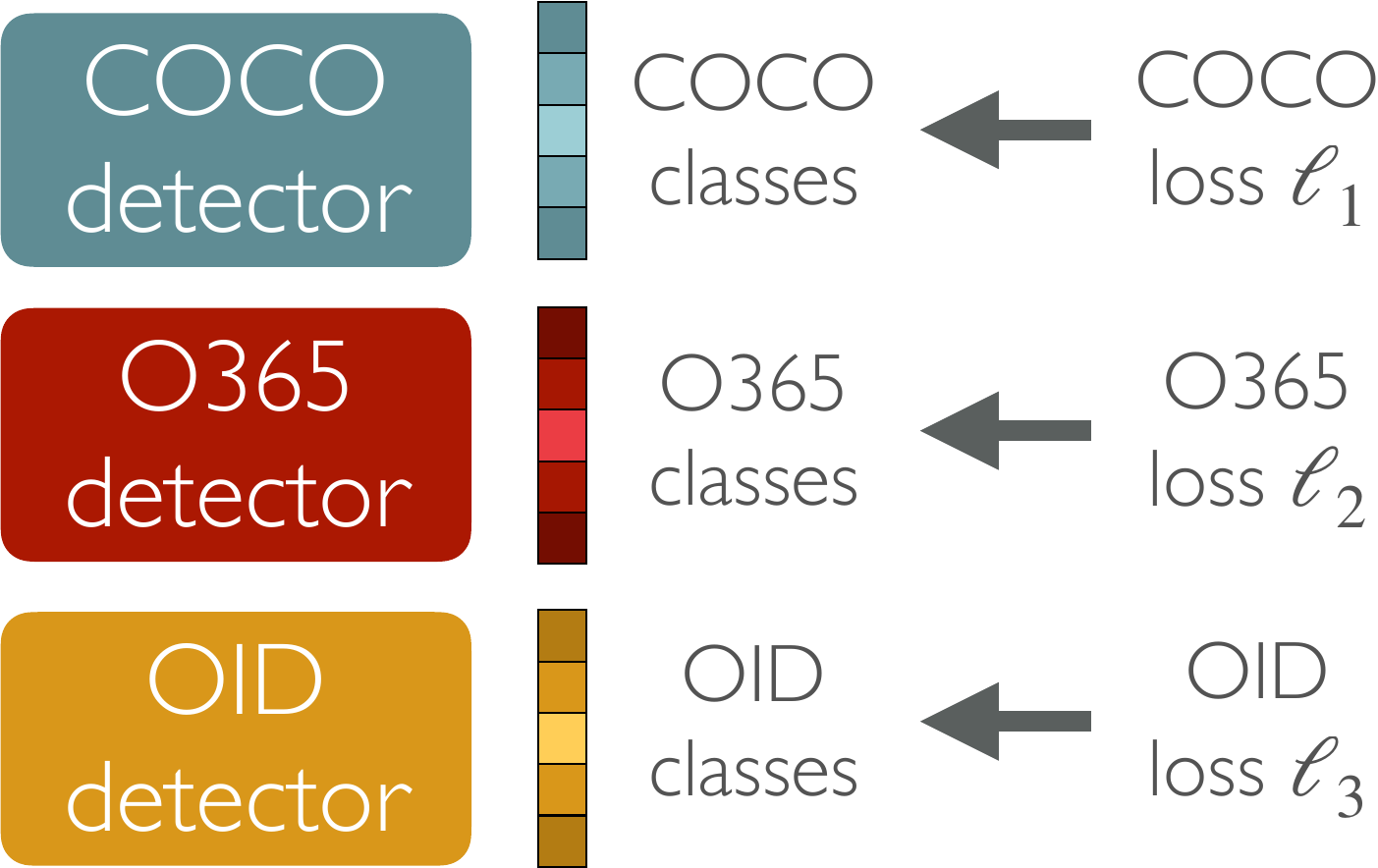}
    \caption{Single-dataset detectors}
    \label{fig:overview_a}
    \end{subfigure}\hfill
    \begin{subfigure}{0.32\textwidth}
    \includegraphics[page=4, width=\linewidth]{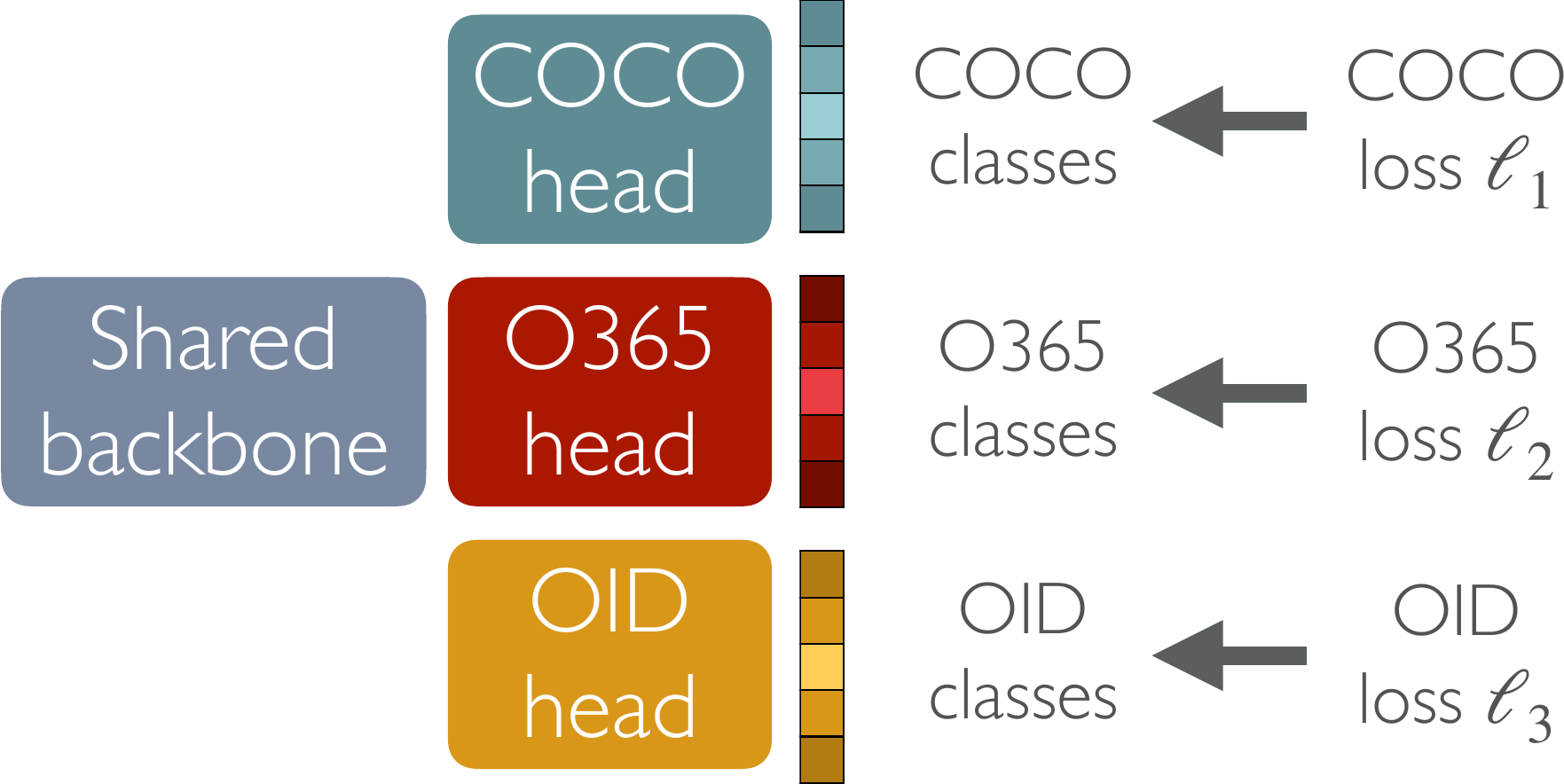}
    \caption{Partitioned detector}
    \label{fig:overview_b}
    \end{subfigure}\hfill
    \begin{subfigure}{0.29\textwidth}
    \includegraphics[page=5, width=\linewidth]{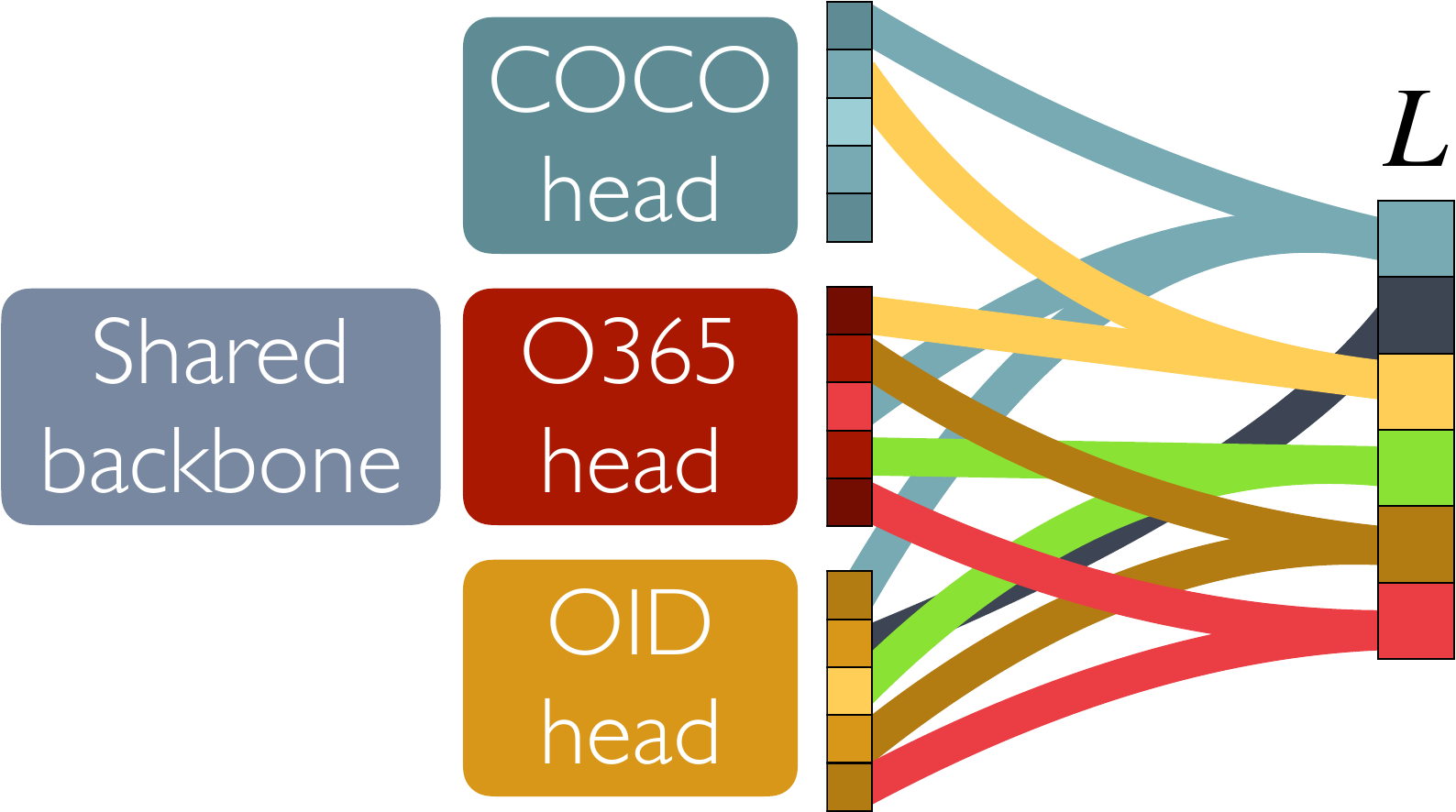}
    \caption{Unified detector}
    \label{fig:overview_c}
    \end{subfigure}
    \vspace{-2mm}
    \caption{
    Standard detectors (a) are trained on one dataset with a dataset-specific loss. We train a single partitioned detector (b) on multiple datasets with shared backbone, dataset-specific outputs and loss. Finally, we unify the outputs of the partitioned detector in a common taxonomy completely automatically (c).
    }
    \label{fig:overview}
    \vspace{-5mm}
\end{figure*}

\lblsec{unified}
\lblsec{labelspace}
\label{sec:learning}

Consider multiple datasets, each with its own label space $L_1, L_2, \ldots$.
Our goal is to jointly learn a common label space $L$ for all datasets, and define a mapping between this common label space and dataset-specific labels ${\T_k : L \to L_k}$.
Mathematically, $\T_k \in \{0,1\}^{|L_k|\times |L|}$ is a Boolean linear transformation.
In this work, we only consider direct mappings.
Each joint label $c \in L$ maps to at most one dataset-specific label $\hat c \in L_k$: $\T_k^\top \vec 1 \le \vec 1$.
I.e., no dataset contains duplicated classes itself.
Also, each dataset-specific label matches to exactly one joint label: $\T_k \vec 1 = \vec 1$.
In particular, we do not hierarchically relate concepts across datasets.
When there are different label granularities, we keep them all in our label-space, and expect to predict all of them\footnote{This follows the official evaluation protocol of OpenImages~\cite{OpenImages}.}.

Given a set of partitioned detector outputs ${d_i^1 \in \mathbb{R}^{|L_1|}, d_i^2 \in \mathbb{R}^{|L_2|}, \ldots}$ for a bounding box $b_i$, we build a joint detection score $d_i$ by simply averaging the outputs of common classes:
\begin{equation}
 d_i = \frac{\sum_k \T_k^\top \vec d^k}{\sum_k \T_k^\top \vec 1},
\end{equation}
where the division is elementwise.
Figure~\ref{fig:overview_c} provides an overview.
From this joint detector, we recover dataset-specific outputs $\tilde d_i^k = \T_k d_i$.
Our goal is to find a set of mappings $\T^\top = \left[\T_1^\top \ldots, \T_N^\top\right]$ and implicitly define a joint label-space $L$ such that the joint classifier does not degrade in performance.

Simple baselines include hand-designed mappings $\T$ and label spaces $L$~\cite{MSeg_2020_CVPR,Zhao_UniDet_ECCV20}, or language-based merging.
One issue with these techniques is that word labels are ambiguous.
Instead, we let the data speak and optimize a label space automatically based on correlations in the firings of 
a pre-trained partitioned detector
on different images, which is a proxy for perceptual similarity.

For a specific output class $c$, let $\mathcal{L}_c$ be a loss function that measures the quality of the merged label space $d_i$ and its re-projections $\hat d_i^k$ compared to the original disjoint label-space $d_i^k$ on a single box $i$.
Let $D^k = [d_1^k, d_2^k, \ldots]$ be the outputs of the partitioned detection head for dataset $\mathcal{D}_k$.
Let $D = \frac{\sum_k \T_k^\top D^k}{\sum_k \T_k^\top \vec 1}$ be the merged detection scores, and $\tilde D^k = \T_k D$ be the reprojection.
Our goal is to optimize this loss over all detector outputs given the Boolean constraints on our mapping
\begin{align}
&\text{minimize}_{L, \T} && E_{\mathcal{D}_k}\left[\sum_{c \in L_k} \mathcal{L}_c(D^k_c, \tilde D^k_c)\right] + \lambda |L|\label{objective_complicated}\\
&\text{subject to} && \T_k \vec 1 = \vec 1 \text{\quad and \quad}\T_k^\top\vec 1 \le \vec 1 \qquad \forall_{k}\notag.
\end{align}
The cardinality penalty $\lambda |L|$ encourages a small and compact label space.
A factorization of the loss $\mathcal{L}_c$ over the output space $c\in L_k$ may seem restrictive.
However, it does include the most common loss functions in detection: score distortion and Average Precision (AP).
Section~\ref{sec:cost} discusses the exact loss functions used in our optimization.

Objective~\ref{objective_complicated} mixes combinatorial optimization over $L$ with a 0-1 integer program over $\T$.
However, there is a simple reparametrization that lends itself to efficient optimization.

First, observe that the label set $L$ simply corresponds to the number of columns in $\T$.
Furthermore, we merge at most one label per dataset $\T_k^{\top} \vec 1 \le \vec 1$.
Hence, for each dataset $\mathcal{D}_k$ a column $\T_k(c) \in \mathbb{T}_k$ takes one of $|\hat L_k|+1$ values: $\mathbb{T}_k = \{\vec 0, \vec 1_{1}, \vec 1_{2}, \ldots\}$, where $\vec 1_{i} \in \{0,1\}^{|L_k|}$ is an indicator vector of the $i$-th element.
Each column $\T(c) \in \mathbb{T}$ then only chooses from a small set of potential values $\mathbb{T} = \mathbb{T}_1 \times \mathbb{T}_2 \times \ldots$, where $\times$ represents the Cartesian product.
Instead of optimizing over the label set $L$ and transformation $\T$ directly, we instead use combinatorial optimization over the potential column values of $\vec t \in \mathbb{T}$.
Let $x_{\vec t} \in \{0,1\}$ be the indicator of combination $\vec t \in \mathbb{T}$.
$x_{\vec t} = 1$ means we apply the class combination specified by $\vec t$, and otherwise not.
In this formulation, the constraint $\T_k \vec 1 = \vec 1 \forall_k$ translates to $\sum_{\vec t\in\mathbb{T} | \vec t(c) = 1}x_{\vec t} = 1$ for all dataset-specific labels $c$.
Furthermore, the objective of the optimization simplifies to
\begin{equation}
\sum_{\vec t \in \mathbb{T}} x_{\vec t} 
\underbrace{E_{\mathcal{D}_k}\left[\sum_{c\in L_k | \vec t(c) = 1}\mathcal{L}_c(D^k_c, \tilde D^k_c)\right]}_{c_{\vec t}} + \lambda \sum_{\vec t \in \mathbb{T}} x_{\vec t}.
\end{equation}
Crucially, the merge cost $c_{\vec t}$ can be precomputed for any subset of labels $\vec t$.
This leads to a compact integer linear programming formulation of objective~\ref{objective_complicated}:
\begin{align}
&\text{minimize}_{x} & \sum_{\vec t \in \mathbb{T}}& x_{\vec t} \left(c_{\vec t} + \lambda\right) \nonumber\\
&\text{subject to} & \sum_{\vec t\in\mathbb{T} | \vec t_{c} = 1}&x_{\vec t} = 1 \qquad \forall_{c}&
\end{align}
For two datasets, the above objective is equivalent to a weighted bipartite matching.
For a higher number of datasets, it reduces to weighted graph matching and is NP-hard, but is practically solvable with integer linear programming~\cite{cylp}.

One drawback of the combinatorial reformulation is that the set of potential combinations $\mathbb{T}$ grows exponentially in the datasets used: $|\mathbb{T}| = O(|\hat L_1| |\hat L_2| |\hat L_3| \ldots)$.
However, most merges $\vec t \in \mathbb{T}$ are bad and incur a large merge cost $c_{\vec t}$.
The supplementary material presents a linear-time greedy enumeration algorithm for low-cost merges, with a pruning hyper-parameter $\tau$.
Considering only low-cost matches, standard integer linear programming solvers find an optimal solution within seconds for all label spaces we tried, even for $|L| > 600$ and up to 6 datasets.

\subsection{Loss functions}
\lblsec{merge_loss}

The loss function in our constrained objective~\ref{objective_complicated} is quite general and captures a wide range of commonly used losses.
We highlight two: an unsupervised objective based on the distortion between partitioned and unified outputs, and Average Precision (AP) on a validation set.

\paragraph{Distortion} measures the difference in detection scores between partitioned and unified detectors:
\begin{equation}
\mathcal{L}_c^\mathrm{dist}(D_c^k, \tilde D_c^k) = \left(D_c^k-\tilde D_c^k\right)^2\label{eq:dist}.
\end{equation}
A drawback of this distortion measure is that it does not take task performance into consideration when optimizing the joint label space.

\paragraph{Average Precision.} Given a reprojected dataset-specific output $\tilde D_c^k$, we can measure the average precision $\mathrm{AP}_{c}(\tilde D_c^k)$ of each output class $c$ on the validation set of $\mathcal{D}_k$.
Our loss measures the improvement in AP:
\begin{equation}
\mathcal{L}_c^\mathrm{AP}(D_c^k, \tilde D_c^k) = \frac{1}{|L_k|}\left(\mathrm{AP}_c(D_c^k) - \mathrm{AP}_c(\tilde D_c^k)\right)\label{eq:map}.
\end{equation}
The AP computation is computationally quite expensive.
We will provide an optimized joint evaluation in our code.

These two loss functions allow us to train a partitioned detector and merge its output space after training, either maximizing the original evaluation metric (AP) or minimizing the change incurred by the unification.

\label{sec:cost}

\section{Experiments}
\lblsec{results}

Our goal is to facilitate the training of a single model that performs well across datasets.
In this section, we first introduce our dataset setup and implementation details.
In \refsec{baseline}, we analyze our key design choices for a partitioned detector baseline.
In \refsec{unified}, we evaluate our unified detector and our unified label space learning algorithm.
We further evaluate the unified detector in new test datasets in a cross-dataset evaluation (\refsec{zeroshot}) without any training on the test domain.

\par \noindent \textbf{Datasets.}
Our main training datasets are adopted from the Robust Vision Challenge (RVC)\footnote{http://www.robustvision.net}.
These are four large datasets for object detection: COCO~\cite{lin2014microsoft}, OpenImages~\cite{OpenImages}, Objects365~\cite{shao2019objects365}, and optionally Mapillary~\cite{MVD2017}.
To evaluate the generalization ability of the models,
we follow MSeg~\cite{MSeg_2020_CVPR} to set up a ross-dataset evaluation protocol: we evaluate models on new test dataset \emph{without training on them}.
Specifically, we test on VIPER~\cite{richter2017playing}, Cityscapes~\cite{cordts2016cityscapes}, ScanNet~\cite{dai2017scannet}, WildDash~\cite{zendel2018wilddash}, KITTI~\cite{Geiger2012CVPR}, 
Pascal VOC~\cite{egwwz-pvocc-10}, and CrowdHuman~\cite{shao2018crowdhuman}.
A detailed description of all datasets is contained in the supplement.
In our main evaluation, we use large and general datasets: COCO, Objects365, and OpenImages. 
Mapillary is relatively small and is specific to traffic scenes; we only add it for the RVC and cross-dataset experiments.

For each dataset, we use its official evaluation metric: 
for COCO, Objects365, and Mapillary, we use mAP at IoU thresholds 0.5 to 0.95. 
For OpenImages, we use the official modified mAP@0.5 that excludes unlabeled classes and enforces hierarchical labels~\cite{OpenImages}.
For the small datasets in cross-dataset evaluation, we use mAP at IoU threshold 0.5 for consistency with PascalVOC~\cite{egwwz-pvocc-10}.

\par \noindent \textbf{Implementation details.}
We use the CascadeRCNN detector~\cite{cai2018cascade} with a shared region proposal network (RPN) across datasets.
We evaluate two models in our experiments:
a partitioned detector (i.e., detector with dataset-specific output heads) and a unified detector.
For the partitioned detector, the last classification layers of all cascade stages are split between datasets.
The unified detector uses CascadeRCNN~\cite{cai2018cascade} as is.

Our implementation is based on Detectron2~\cite{wu2019detectron2}.
We adopt most of the default hyper-parameters for training.
We use the standard data augmentation, including random flip and scaling of the short edge in the range $[640, 800]$.
We use SGD with base learning rate 0.01 and batch size 16 over 8 GPUs.
We use ResNet50~\cite{he2016deep} as the backbone in our controlled experiments unless specified otherwise.
We use a $2\times$ training schedule (180k iterations with learning rate dropped at the 120k and 160k iterations)~\cite{wu2019detectron2} in most experiments unless specified otherwise, regardless of the training data size.

\subsection{Multi-dataset detection}
\lblsec{baseline}

\begin{table}[t]
\centering
\begin{tabular}{@{}l@{}c@{\ }c@{\ }c@{\ }c@{}}
\toprule
 & COCO & O365 & OImg & \emph{mean} \\
\midrule
Simple merge~\cite{wang2019towards}\! & 34.2 & 14.6 & 50.8 & 33.2 \\
w/ uniform dataset sampling & 41.1 & 16.5 & 46.0 & 34.5 \\
w/ class-aware sampling  & 35.3 & 18.5 & 61.8 & 38.5 \\
w/ dataset+class-aware sampling & {\bf 41.8} & 20.3 & 60.0 & 40.6 \\
\midrule
Partitioned detector (ours) & {\bf 41.8} & {\bf 20.6} & {\bf 62.7} & {\bf 41.7} \\
\bottomrule
\end{tabular}
\caption{
\textbf{Effectiveness of our multi-dataset training strategies}.
We start with a simple merging of datsets~\cite{wang2019towards}, then add a uniform sampling of images between different training datasets (second row), class-aware sampling within Objects365 and OpenImages (third row), and both sampling strategies (fourth row).
Our partitioned detector combines these sampling strategies with a dataset-specific loss (last row). 
}
\lbltbl{sampling}
\vspace{-5mm}
\end{table}

\begin{table*}[!b]
\vspace{-3mm}
\centering
\begin{tabular}{@{}l@{\ \ }c@{\ \ }c@{\ \ \ }c@{\ \ }c@{\ \ }c@{\ \ }c@{\ \ }c@{ \ \ }c@{\ \ }c@{}}
\toprule
 & \multicolumn{3}{c}{$2\times$} & \multicolumn{3}{c}{$6\times$} & \multicolumn{3}{c}{$8\times$} \\
 & COCO & Objects365 & OImg.  & COCO & Objects365 & OImg.  & COCO & Objects365 & OImg. \\
\cmidrule(r){1-1}
\cmidrule(r){2-4}
\cmidrule(r){5-7}
\cmidrule(r){8-10}
Partitioned detector & {\bf 41.8} & 20.6 & 62.7 & {\bf 44.6} & 23.6 & 64.8 & {\bf 45.5} & 24.6 & {\bf 66.0} \\
\cmidrule(r){1-1}
\cmidrule(r){2-4}
\cmidrule(r){5-7}
\cmidrule(r){8-10}
COCO & 41.5 & - & - & 42.5 & - & - & 42.5 & - & - \\
Objects365 & - & {\bf 23.8} & - & - & {\bf 25.0} & - & - & {\bf 24.9} & - \\
OpenImages & - & - & {\bf 64.6} & - & - & {\bf 65.4} & - & - & 65.7 \\
\bottomrule
\end{tabular}
\vspace{-3mm}
\caption{\textbf{Dateset-specific vs partitioned detectors.} We show validation mAP of our partitioned model and the three dataset-specific models under different training schedules. The performance of a partitioned model matches dataset-specific models on long schedules.}
\label{table:schedule}
\lbltbl{schedule}
\end{table*}

\begin{figure*}[!t]
\centering
   \includegraphics[page=1,width=0.95\linewidth]{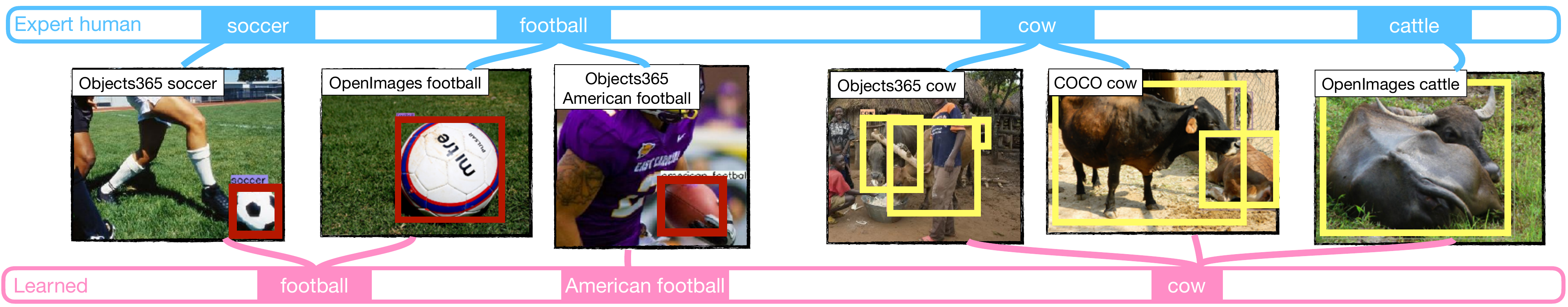}
   \includegraphics[page=2,width=0.95\linewidth]{figs/UniDet_examples4.2.pdf}
   \caption{
   \textbf{Sampled results of the learned unified label space}.
   We show example differences between an expert-designed label space provided as part of the Robust Vision Challenge (top of each row, blue) and our learned label space (bottom of each row, pink). Our learned label space captures detailed visual differences. Zoom in for details.}
\label{fig:labelspace}
\vspace{-5mm}
\end{figure*}

We first evaluate the partitioned detector.
We use dataset-specific outputs and do not merge classes between different datasets. 
During evaluation, we assume the target dataset is known and only look at the corresponding output head.
As discussed in \refsec{splithead}, our baseline highlights two basic components: uniform sampling of images between datasets and dataset-specific training objective.
For these experiments we distinguish between modifications of the objective that merely sample data differently within each dataset (e.g.\ class-aware sampling), and changes to the loss functions (e.g.\ hierarchical losses).

We start from the baseline of \cite{wang2019towards,wu2019detectron2}.
They simply collect all data from all datasets and train with a common loss.
As is shown in~\reftbl{sampling}, this biases the model to large datasets (OpenImages) and yields low performance for relatively small datasets (COCO).
Sampling datasets uniformly (second row) trades the performance on smaller datasets with large datasets, and overall improves performance.
On the other hand, both OpenImages and Objects365 are long-tailed and best train with advanced inter-dataset sampling strategy~\cite{shen2016relay,peng2020large}, namely class-aware sampling.
Class-aware sampling significantly improves accuracy on OpenImages and Objects365.
Combining the uniform dataset sampling and the intra-dataset class-aware sampling gives a further boost.
Finally, OpenImages~\cite{OpenImages} requires predicting a label hierarchy. 
For example, it requires predicting ``vehicle'' \emph{and} ``car'' for all cars. 
This breaks the default cross-entropy loss that assumes exclusive class labels per object.
We instead use a dedicated hierarchy-aware sigmoid cross-entropy loss for OpenImages~\cite{OpenImages}.
Specifically, for an annotated class label in OpenImages, we set all its parent classes as positives and ignore the losses over its descendant classes.
Our partitioned detector combines both sampling strategies and the dataset-specific loss.
The hierarchy-aware loss yields a significant $+2.7$mAP improvement on OpenImages alone, and does not degrades other datasets.

\vspace{2mm}
\par \noindent \textbf{Dateset-specific vs.\ partitioned detectors.}
In our partitioned detector, training on multiple datasets resembles training separate individual models but with a shared detector.
\reftbl{schedule} compares training a partitioned detector on all datasets with dataset-specific models.
We compare detectors under different training schedules ($n\times$ the COCO default schedule).
Each of the three dataset-specific models sees the same number of gradient updates as our partitioned detector.
In a $2\times$ training schedule (180k iterations), single-dataset models generally perform better than a partitioned model, 
as each dataset is only trained for a $\frac{1}{3}\times$ schedule in the partitioned model.
At a $6\times$ schedule, the partitioned detector starts to match dataset-specific models, and outperforms $2\times$ dataset-specific models under the same total iterations.
In a $8\times$ schedule, all models converge.
The partitioned detector surpasses the single-dataset model on COCO, and matches OpenImages and Objects365 models.

\subsection{Unified multi-dataset detection}

\lblsec{unified}

Next, we evaluate different ways to unify the label space.

\par \noindent \textbf{Unified label space}
We run our label space learning algorithm from \refsec{labelspace} based on the output of a partitioned detector with a ResNeSt backbone~\cite{zhang2020resnest} trained on COCO, Objects365, and OpenImages, with a total of $945$ disjoint classes. 
The hyperparameters are $\lambda = 0.5$ and $\tau = 0.25$.
The optimization ends up with a unified label space with cardinality $|L|=701$.
we compare our automated data-driven unification to human and language-based baselines.
We use the official manually-crafted RVC taxonomy as the human expert baseline\footnote{\scriptsize{\url{https://github.com/ozendelait/rvc_devkit/blob/master/objdet/obj_det_mapping.csv}}}.

Over two-thirds of our learned label space agrees with the human expert. \reffig{labelspace} highlights some of the differences.
Our unification successfully groups similar concepts with different descriptions (``Cow'' and ``Cattle''), and is not distracted by spurious linguistic matches (``American football'' and ``football'').
Interestingly, the learned label space splits the ``oven'' classes from COCO, Objects365, and OpenImages, even though they share the same word.
A visual examination reveals that they are visually dissimilar due to different underlying definitions of the ``oven'' concept in the different datasets: COCO ovens include the cooktop, OpenImages ovens include the control panel, and Objects365 ovens are just the front door. Our data-driven taxonomy reconciliation is able to detect such distinctions, which are missed by word-level approaches.

\begin{table}[!t]
\centering
\centering
\begin{tabular}{@{}l@{\ \ }c@{\ \ }c@{\ }c@{\ }c@{\ }c@{}}
\toprule
 & $|L|$ & COCO & O365 & OImg. & \emph{mean} \\
\midrule
GloVe embedding & 696 & 41.6{\tiny $\pm$0.00} & 20.3{\tiny $\pm$0.12} & 62.4{\tiny $\pm$0.06} & 41.4{\tiny $\pm$0.05} \\
Learned, distortion & 682 & 41.6{\tiny $\pm$0.15} &  20.7{\tiny $\pm$0.06} & 62.6{\tiny $\pm$0.06} & 41.7{\tiny $\pm$0.09} \\
Learned, AP (ours) & 701 & \bf{41.9}{\tiny $\pm$0.10} &  \bf{20.8}{\tiny $\pm$0.10}& \bf{63.0}{\tiny $\pm$0.21}& \bf{41.9}{\tiny $\pm$0.02}\\
\midrule
Expert human & 659 & 41.5{\tiny $\pm$0.06}& 20.7{\tiny $\pm$0.06}& 62.6{\tiny $\pm$0.06} & 41.6{\tiny $\pm$0.04}\\
\bottomrule
\end{tabular}
\vspace{-3mm}
\caption{
\textbf{Evaluation of unified label spaces.} 
We show label space size ($|L|$) and mAP on the validation sets of the training domains. We compare to a language-based baseline (GloVe) and a manual unification by a human expert. Each model is a ResNet50 CascadeRCNN trained in a $2\times$ schedule. We show the mean and standard deviation based on 3 repeated runs. Our learned label space works better than the language and the human counterparts.}
\label{table:space}
\vspace{-3mm}
\end{table}

We next quantitatively compare our learned label space with alternatives.
For each label space, we retrain a multi-dataset detector with that label space.
During training, as with our partitioned model, we only apply training losses to the classes that are annotated in the source dataset.
We compare our learned label space to a ``best effort'' human baseline and a language-based baseline.
For the language-based baseline, we replace the cost measurement defined in Section~\ref{sec:cost} with the cosine distance between the GloVe word embeddings~\cite{pennington2014glove}, and run the same integer linear program.
Table~\ref{table:space} shows the results.
We repeat the training for three runs with different random seeds and report the mean and standard deviation. 
The four label spaces agree on most classes and the overall mAP is thus close.
Our automatically constructed label space consistently outperforms the human expert baseline, with a healthy $0.3$ mAP margin on average.
The improvement appears statistically stable under multiple training runs.
Notably, the relative improvement of our model over the expert is larger than the expert's improvement over the language-based baseline.

\begin{table}[!t]
\centering
\centering
\begin{tabular}{@{}l@{\ \ }l@{\ \ \ \ }c@{\ \ \ \ }c@{\ \ \ \ }c@{\ \ \ \ }c@{\ \ \ \ }c@{}}
\toprule
 $\lambda$ & $\tau$ & $|L|$ & COCO & O365 & Oimg. & mean\\
\midrule
0.1 & 0.25 & 700 & \bf 41.9 & 20.6 & 62.9 & 41.8 \\
0.5* & 0.25* & 701 & \bf 41.9 & 20.8 & \bf 63.0 & \bf 41.9 \\
1.0 & 0.25 & 703 & \bf 41.9 & \bf 20.9 & \bf 63.0 & \bf 41.9 \\
0.5 & 0.2 & 668 & 41.6 & 20.7 & 62.9 & 41.7 \\
0.5 & 0.3 & 728 & 41.8 & \bf 20.9 & 62.9 & \bf 41.9 \\
\bottomrule
\end{tabular}
\vspace{-3mm}
\caption{\textbf{Hyper-parameter choices}.
We change $\lambda$ and $\tau$ of the label space learning algorithm. We show the size of the resulting label space and the mAP on 3 datasets. *: the default option. The pruning threshold $\tau$ impacts the label space size, but not mAP.}
\label{table:hyper}
\vspace{-5mm}
\end{table}

\noindent\textbf{Hyper-parameter choices.} 
Table~\ref{table:hyper} ablates the hyper-parameters $\lambda$ and $\tau$ of the label space learning algorithm (\refsec{labelspace}). 
Our algorithm is robust to the cardinality penalty factor $\lambda$. 
Varying the cardinality penalty $\lambda$ from $0.1$ to $1.0$ only affects the size of the label space by $3$.
The pruning threshold $\tau$ has a larger impact on the label space size, but not the final performance.
We use $\lambda=0.5$ and $\tau=0.25$ for a good balance between the label space size and overage performance.

\noindent\textbf{Unified vs.\ partitioned detectors.}
\label{sec:trainingset}
We next compare unified detectors with and without retraining using the joint taxonomy, a partitioned detector, and an ensemble of dataset-specific detectors.
The partitioned detector and the ensemble need to know the target domain at test time, while the unified models do not.
This means that the unified models can be deployed without any modification in new domains, while the alternatives must know which domain they are in.
Table~\ref{table:trainingset} shows the results.
A partitioned detector outperforms a dataset-specific ensemble under the same conditions (Table~\ref{table:trainingset} bottom), especially on the ``small'' COCO dataset.
An offline unification loses some accuracy, but this is regained when retraining the model under the unified taxonomy (Table~\ref{table:trainingset} top).
Crucially, the unified models do not need to know what domain they are in at test time.

\begin{table}[!t]
\centering
\begin{tabular}{@{}l@{}c@{\ \ \ \ }c@{\ \ \ \ }c@{\ \ \ \ }c@{}}
\toprule
 & COCO & O365 & OImg. & \emph{mean} \\
\midrule
Unified (naive merge) & 44.4 & 23.6 & 65.3 & 44.4\\
Unified (retrained)  & 45.4 & 24.4 & 66.0 & 45.3 \\
\midrule
Partitioned (oracle) & 45.5 & 24.6 & 66.0 & 45.4 \\
Ensemble (oracle) & 42.5 & 24.9 & 65.7 & 44.4 \\
\bottomrule
\end{tabular}
\vspace{-3mm}
\caption{
\textbf{Unified vs.\ partitioned detectors}.
We show validation mAP on training domains for a unified detector directly from merging partitioned detector weights (top), the same detector retrained on the joint taxonomy (second), a partitioned detector knowing the target domain (thrid), and an ensemble of three dataset-specific detectors (bottom).
The bottom two rows require a known test dataset source and the top two rows do not.
All models use a ResNet-50 CascadeRCNN trained in an $8\times$ schedule.}
\label{table:trainingset}
\vspace{-5mm}
\end{table}

\subsection{Cross-dataset evaluation}

\label{sec:zeroshot}
\lblsec{zeroshot}

\begin{table*}[!t]
\centering
\begin{tabular}{@{}l@{\ \ \ \ }l@{\ \ \ \ }c@{\ \ \ \ }c@{\ \ \ \ }c@{\ \ \ \ }c@{\ \ \ \ }c@{\ \ \ \ }c@{\ \ \ \ }c@{\ \ \ \ }c@{\ \ \ \ }c@{\ \ \ \ }c@{\ \ \ \ }c@{}}
\toprule
\rowNumber{\#} & & VOC & VIPER & Cityscapes  & ScanNet & WildDash & CrowdH. & KITTI & \emph{mean}\\
\midrule
\rowNumber{1} & COCO & 80.0 & 13.9 & 39.6  & 17.4 & 25.9 & \bf{73.9} & 30.5 & 40.2\\
\rowNumber{2} & Objects365 & 71.9 & 20.7 & 43.4  & 24.9 & 27.6 & 71.8 & 32.2 & 41.8 \\
\rowNumber{3} & OpenImages & 64.4 & 10.4 & 29.8 & 24.2 & 20.3 & 66.7 & 21.8 &  33.9 \\
\rowNumber{4} & Mapillary & 11.4 & 15.2 & 44.7 & 0.0 & 23.4 & 49.3 & 37.8 & 26.0 \\
\rowNumber{5} & Ensemble & 79.7 & 16.8 & 46.0 & 30.1 & 32.1 & \bf{73.9} & 34.3 & 44.7 \\
\rowNumber{6} & Partitioned & {\bf 83.1} & 20.9 & 48.4 & {\bf 32.2} & 34.4 & 70.0 & 38.9 & 46.8 \\
\rowNumber{7} & Unified (retrained) & {\bf 82.9} & {\bf 21.3} & {\bf 52.6} & 29.8 & {\bf 34.7} & 70.7 & {\bf 39.9} & {\bf 47.3} \\
\midrule
\rowNumber{8} & \textcolor{gray}{Dataset-specific} & \textcolor{gray}{80.3} & \textcolor{gray}{31.8} & \textcolor{gray}{54.6} & \textcolor{gray}{44.7} & \textcolor{gray}{-} & \textcolor{gray}{80.0} & \textcolor{gray}{-} & \textcolor{gray}{-} \\
\bottomrule
\end{tabular}
\vspace{-2mm}
\caption{
\textbf{Cross-dataset evaluation}.
We show mAP50 on the validation sets of datasets that were not seen during training. We compare models trained on each single training dataset (Rows \rowNumber{1-4}), the ensemble of the 4 single dataset models (row \rowNumber{5}), a partitioned detector (row \rowNumber{6}), and the unified detector with our learned unified label space (row \rowNumber{7}).
For reference, we show the {\textcolor{gray}{``oracle''}} models that are trained on the training set of each test dataset on row \rowNumber{8}. The columns refer to test datasets. Each model is a ResNet-50 CascadeRCNN trained until converge or at most an $8\times$ schedule.}
\label{table:zeroshot}
\vspace{-5mm}
\end{table*}

We evaluate the generalization ability of object detectors by evaluating them in new test domains not seen during training.
In this setting, we do not assume to know the test classes ahead of time.
To allow for a fair and unbiased evaluation, we use a simple language-based matching to find the test-to-train label correspondence.
Specifically, we calculate the GloVe~\cite{pennington2014glove} word embedding distances between each test label and the training label, and match the test label to its closest training label.
If multiple training labels match, we break ties in a fixed order: COCO, Objects365, OpenImages, and Mapillary\footnote{We also tried evaluating under different orders, and find the listed order to perform best for all methods.}.

We compare both our multi-dataset models (partitioned or unified) to single-dataset models.
We use all four RVC training sets to train the multi-dataset models.
Specifically, we start from a $6\times$ schedule model trained on the three large datasets, and add Mapillary~\cite{MVD2017} in a $2\times$ fine-tuning schedule with $10 \times$ smaller learning rate.
We compare all models under the same schedule
\footnote{except for the Mapillary model, for which a $2\times$ schedule performs better than longer schedules.}, 
hyperparameters, and detection models.
In addition, we also compare to the ensemble of the four single-dataset models trained analogously to the partitioned model.
For reference, we also show the performance of detectors trained on the training set of each test dataset.
This serves as an oracle ``upper bound'' that has seen the test domain and label space.
Note that KITTI and WildDash are small and do not have a validation set. We thus evaluate on the training set and do not provide the oracle model.

Table~\ref{table:zeroshot} shows the results.
The COCO model exhibits reasonable performances of some test datasets, such as Pascal VOC and CrowdHuman.
However, its performance is less than satisfactory on datasets such as ScanNet, whose label space differs significantly from COCO.
Training on the more diverse Objects365 dataset yields higher accuracy in the indoor domain, but loses ground on VOC and CrowdHuman, which are more similar to COCO.
Training on all datasets, either with a partitioned detector (row \rowNumber{6}) or a unified one (row \rowNumber{7}) yields generally good performance on all test datasets.
Notably, both our detectors perform better than the ensemble of the 4 single dataset models (row \rowNumber{5}), showing that the multi-dataset models learned more general features.
On Pascal VOC, both multi-dataset models outperform the VOC-trained upper-bound without seeing VOC training images. 
Our unified model outperforms the partitioned detector overall and operates on a unified taxonomy.

\subsection{Scale up to large models}

\begin{table}[t]
\centering
\begin{tabular}{@{}l@{}c@{\ \ }c@{\ }c@{\ }c@{}}
\toprule
 & COCO & OImg. & Mapillary & O365\\
\midrule
Ours & \bf{52.9} & \bf{60.6}/\bf{56.8} & \bf{25.3} & \bf{33.7} \\
ResNeSt200~\cite{zhang2020resnest} & 50.9 & - & - & - \\
TSD~\cite{song2020revisiting} & - & 60.5/- & - & - \\
CACascade RCNN~\cite{gao2019objects365} & - & - & - & 31.6 \\
\bottomrule
\end{tabular}
\vspace{-3mm}
\caption{\textbf{Scale up to large models}. We show results on COCO test-challenge set, OpenImages challenge 2019 test sets (public test set/ private test set), Mapillary test set, and Objects365 validation set. Top row: our detector with a ResNeSt200 backbone. 2-4 rows: state-of-the-art single-dataset models with comparable backbones (without model ensembles or test-time augmentation).
}
\label{table:challenge}
\vspace{-5mm}
\end{table}

\lblsec{rvcchallenge}

Next, we scale up our unified detector with a large backbone to develop a ready-to-deploy object detector.
We used a ResNeSt200 backbone~\cite{zhang2020resnest} and followed the same training procedure as in Section~\ref{sec:trainingset} with an $8\times$ schedule.
The training took $\sim$16 days on a server with 8 Quadro RTX 6000 GPUs.
Table~\ref{table:challenge} shows our \emph{single} model achives $52.9$ mAP on COCO, $60.6$ mAP on OpenImages, and $33.7$ mAP on Objects 365.
We compare to state-of-the-art results with comparable baselines on each individual dataset.
On COCO, our result improves the COCO-only ResNeSt200~\cite{zhang2020resnest} model, by $2$ mAP with the same detector, thanks to our ability to train with more data.
On OpenImages, our result matches the best single model in the OpenImages 2019 Challenge, TSD~\cite{song2020revisiting}, with a comparable backbone (SENet154-DCN~\cite{hu2018squeeze} of TSD). 
On Objects365, we outperform the 2019 Object365 detection challenge winner~\cite{gao2019objects365} by $2$ mAP points.

\section{Conclusion}
We presented a simple recipe for training a single object detector across multiple datasets
and a formulation to automatically construct a unified taxonomy.
Our resulting detector can be deployed in new domains without additional knowledge.
We hope our model makes object detection more accessible to general users.

\noindent\textbf{Limitations.} Our label space learning algorithm currently uses only visual cues,
integrating language cues as auxiliary information may further improve the performance.
Our formulation currently does not consider label hierarchies, and the resulting label space treats COCO person and OpenImages boy as two independent classes.
We leave incorporating label hierarchies as exciting future work. 
{\small
\par \noindent \textbf{Acknowledgments.}
This material is based upon work supported by the National Science Foundation under Grant No. IIS-1845485 and IIS-2006820.
Xingyi is supported by a Facebook Fellowship.
}

{\small
\bibliographystyle{ieee_fullname}
\bibliography{egbib}
}

\appendix

\section{Dataset details}
\label{appendix:datasets}

\begin{table}[!b]
\centering
\begin{tabular}{@{\ \ \ }l@{\ \ \ }c@{\ \ \ }c@{\ \ \ }c}
\toprule
Dataset name & Domain  & \# Cat. & \# Img. \\
\midrule
\textbf{Train \& Validation} \\
\ \ \ \ COCO & Internet images & 80 & 118k \\
\ \ \ \ Objects365 & Internet images & 365 & 600k\\
\ \ \ \ OpenImages & Internet images & 500 & 1.8M \\
\ \ \ \ Mapillary & Traffic & 38 & 18k\\
\textbf{Test} \\
\ \ \ \ ScanNet & Indoor & 20 & 25k \\
\ \ \ \ VIPER & Virtual & 10 & 13k\\
\ \ \ \ Cityscapes & Traffic & 8 & 12k \\
\ \ \ \ WildDash & Traffic & 13 & 4k \\
\ \ \ \ KITTI & Traffic & 8 & 200 \\
\ \ \ \ Pascal VOC & Internet images & 20 & 16k\\
\ \ \ \ CrowdHuman & Internet images & 1 & 15k\\
\bottomrule
\end{tabular}
\vspace{-2mm}
\caption{Datasets we used in training and testing. Top: datasets we used in training and validation, which are from the Robust Vision Challenge. Bottom: datasets we used for zero-shot cross-dataset testing.}
\lbltbl{datasets}
\vspace{-5mm}
\end{table}

~\reftbl{datasets} lists the datasets we used in our experiments.
We use the Robust Vision Challenge\footnote{\url{http://www.robustvision.net}} official release of each dataset.
Specifically, we use the standard 2017 train/ validation split for COCO~\cite{lin2014microsoft}, the Challenge-2019 release of OpenImages~\cite{OpenImages}, and the default version of Objects365~\cite{shao2019objects365} and Mapillary~\cite{MVD2017}.
For ScanNet~\cite{dai2017scannet}, as there is no standard train/ validation split, we use the first $80\%$ scenes (sorted by scene ID) as training and the last $20\%$ scene as validation.
For KITTI~\cite{Geiger2012CVPR}, we used the RVC challenge version that has instance-segmentation version, which contains 200 images.
For WildDash~\cite{zendel2018wilddash}, we use the public version for evaluation, and report standard mAP performance.
We don't consider the negative label metric in the official website.
For CrowdHuman~\cite{shao2018crowdhuman}, we use the visible bounding box annotation, and report mAP instead of the missing rate as the official metric.
We use the official train/ validation split and the official evaluation metrics for VIPER~\cite{richter2017playing}, Cityscapes~\cite{cordts2016cityscapes}, and Pascal VOC~\cite{egwwz-pvocc-10}.
\begin{table*}[!t]
\centering
\begin{tabular}{@{}l@{\ \ }c@{\ \ }c@{\ \ }c@{\ \ }c@{\ \ }c@{\ \ }c@{\ \ }c@{\ \ }c@{\ \ }c@{}}
\toprule
 & COCO & CityScapes & Mapillary  & VIPER & ScanNet & OpenImages & KITTI & WildDash\\
\cmidrule(r){1-1}
\cmidrule(r){2-7}
\cmidrule(r){8-9}
COCO & \bf{35.6} & 19.6 & 3.2 & 8.5 & 5.2 & 7.2 & 15.7 & 8.4\\
CityScapes & 0.0 & 21.5 & 0.8 & 2.3 & 0.0 & 0.0 & 13.0 & 2.4 & \\
Mapillary & 0.6 & 11.7 & \bf{10.6} & 9.0 & 1.2 & 0.0 & 13.4 & 5.4\\
VIPER  & 0.1 & 2.8 & 1.1 & \bf{17.8} & 0.0 & 0.0 & 6.5 & 1.4 &  \\
ScanNet & 0.4 & 0.0 & 0.0 & 0.0& \bf{35.6} & 0.0 & 0.0 & 0.0\\
OpenImages & 12.9 & 9.5 & 1.1 & 3.5 & 1.7 & \bf{52.8} & 7.2 & 4.9\\
\cmidrule(r){1-1}
\cmidrule(r){2-7}
\cmidrule(r){8-9}
Unified (ours) & 24.0 & \bf{28.3} & 8.1 & 16.5 & 28.7 & 41.8 & \bf{16.9} & \bf{11.3}\\
\bottomrule
\end{tabular}
\caption{Instance segmentation performance on six training datasets and two new datasets (KITTI and WildDash). We show mask mAP on the validation set of each dataset.}
\label{table:zeroshot_inst}
\end{table*}

\section{Computation of label space learning algorithm and pruning}
\label{sec:prune}

The size of our optimization problem scales linearly in the number of potential merges $|\mathbb{T}|$, which can grow exponentially in the number of datasets.
To counteract this exponential growth, we only consider sets of classes $$\mathbb{T}^\prime = \left\{\vec t \in \mathbb{T} \bigg| \frac{c_{\vec t}}{|\vec t|-1} \le \tau\right\}.$$
For an aggressive enough threshold $\tau$, the number of potential merges $|\T^\prime|$ remains manageable.
We greedily grow $\T^\prime$ by first enumerating all feasible two-class merges ($|\vec t|=2$), then three-class merges, and so on.
The detailed algorithm diagram is shown in Algorithm~\ref{alg:labelspace}.
The runtime of this greedy algorithm is $O(|\T^\prime|\max_i|\hat L^i|)$.
In practice, the cost computation took a few seconds for the distortion loss function and about 10 minutes for the AP loss (due to the need to repeatedly recompute AP).
The integer programming solver finds the optimal solution within one second in both cases.

\begin{algorithm}[!t]
    \caption{Learning a unified label space}
	\label{alg:labelspace}
	\SetAlgoLined
	\SetKwInOut{Input}{Input} \SetKwInOut{Output}{Output} 
    \Input{$\{{\bf b}_i, \hat{\bf l}_i\}_{i=1}^N$: ground truth bounding boxes and labels for each of the N training datasets\\
      $\{\{\tilde{\bf b}_i^{(j)}, \tilde{\bf l}_i^{(j)}\}_{j=1}^{N}\}_{i=1}^{N}$: predicted bounding boxes with predicted classes in all datasets for each training dataset\\
      $\lambda, \tau$: hyper-parameters for algorithm\\
     }
	\Output{L: unified label space\\
	  $\mathcal{T}$: the transformation from each individual label space to the unified label space}
	// Compute potential merges and merge cost\\
	$\hat L = \bigcup_i \hat L_i$ // Short-hand used to simplify notation\\
	$\mathbb{T}_1 \gets \{(l) | l \in \hat L\}$  // Set of single labels\\
	Compute $c_{\vec t}$ for all single labels $\vec t \in \mathbb{T}$. // 0 for most metrics\\
	\For{$n = 2 \ldots N$}{
	  $\mathbb{T}_n \gets \{\}$\\
	  \For{$\vec t \in \mathbb{T}_{n-1}$} {
    	\For{$l \in \hat L$} {
    	  \If{$l$ and all labels in $\vec t$ are from different datasets}{
    	    compute $c_{\vec t \cup \{l\}}$.\\
    	    \If{$\frac{c_{\vec t \cup \{l\}}}{n-1} \le \tau$}{
    	        Add $\vec t \cup \{l\}$ to $\mathbb{T}_n$.
    	    }
    	  }
    	}
	  }
	}
	$\mathbb{T} \gets \bigcup_{n=1}^N \mathbb{T}_n $ \\
    // Solve the ILP.\\
    $\vec{x} \gets  \text{ILP\_solver}(c, \mathbb{T}, \lambda)$ // Solve equation (8).\\
    Compute $L, \mathcal{T}$ from $\vec{x}$ \\
    \textbf{Return}: $L, \mathcal{T}$
\end{algorithm}

\section{Adding new datasets to a label space}
\label{appendix:expanding}

While we tend to keep the training domains and label space large and comprehensive, it is inevitable in practice that more fine-grained labels or specific testing domains are needed.
Given a learned a unified label space on an existing set of training datasets, we use a simple label space expansion algorithm to allow adding more datasets and labels after the unified detector is trained.

Similar to our unified label space learning algorithm, we run the unified detector on the new training data.
We evaluate the AP between each class in the new dataset annotation and each class in the unified label space.
We merge the new class into the existing class that gives the lowest merge cost (Section. 4.2). 
In our experiments, add Mapillary dataset~\cite{MVD2017} to our label space we using the AP loss.
If the cost is lower than a threshold (AP change $<5$ AP in our implementation).
Otherwise, we append the new class to the unified label space as a single class.

\section{Discussion on label hierarchy}
Different datasets may contain different label granularities for the same concept, 
and there exists label hierarchies inter or intra datasets. 
For example, Objects365~\cite{shao2019objects365} does not have a ``bird'' category, but has more fine-grained bird species like parrot, pigeon, and swan, while most other datasets only annotate ``bird''.
Our label space optimization algorithm automatically handles the hierarchical label space issue:
the fine-grained birds in Objects365 will not merge with COCO birds because this merge introduces many false-positives for the fine-grained birds in Objects365 and yields a large cost.
Our unified label space will contain both the general ``bird'' class and each fine-grained class.
The model trained on the unified label space is expected to predict both the coarse ``bird'' label and the fine-grained label in testing.

\section{Instance segmentation}
\label{appendix:instance}

We further evaluate our label space learning algorithm and unified training framework on instance segmentation. 
We follow the Robust vision challenge set up to use 8 datasets: COCO, OpenImages, Mapillary, ScanNet, VIPER, CityScapes, WildDash and KITTI (the same as ~\reftbl{datasets}, except OpenImages segmentation set has 300 instead of 500 classes.).
Again, we leave WildDash and KITTI as testing only as they are small and similar to CityScapes and Mapillary.
We run our label space learning algorithm (Section. 4) on the remaining six datasets, resulting a unified label space of $358$ classes.
We use CascadeRCNN~\cite{cai2018cascade} with a standard mask head as the detector, and train a $2\times$ schedule with ResNet50.
The dataset-specific models are trained with $1\times$ or $2\times$ schedule depending on their size.

Table.~\ref{table:zeroshot_inst} compares the unified detector to dataset specific models. 
As expected, no single dataset-specific model performs well on all test domains. 
Our unified model performs consistently good on all training datasets.
More importantly, it generalizes the best to the new test datasets (KITTI and WildDash) than any single dataset model.

\end{document}